\documentclass[10pt,twocolumn,letterpaper]{article}
 \usepackage[pagenumbers]{cvpr} 
\usepackage{graphics}
\usepackage{amsmath}
\usepackage{diagbox}
\usepackage{epsfig}
\usepackage{threeparttable}
\usepackage{xspace}
\usepackage{siunitx}
\usepackage{graphicx}
\usepackage{amssymb}
\usepackage{color}
\usepackage{multirow}
\usepackage{float}
\usepackage{amsfonts}
\usepackage{xcolor}
\usepackage{color, colortbl}
\usepackage{bm}
\usepackage{array}
\PassOptionsToPackage{table}{xcolor} 
\definecolor{mygray}{gray}{.90}
\definecolor{reda}{RGB}{202,0,0}
\definecolor{redb}{RGB}{217,148,143}
\definecolor{myyellow}{RGB}{190,144,0}
\definecolor{mygreen}{RGB}{0,136,51}
\definecolor{myblue}{RGB}{0,102,204}
\usepackage{colortbl}

\usepackage{pifont}

\usepackage{epsfig}
\usepackage{graphicx}
\usepackage{amsmath}
\usepackage{amssymb}
\usepackage{enumitem}
\usepackage{tabu}
\usepackage{tabularx}
\usepackage{booktabs}
\usepackage{diagbox}
\usepackage{array}
\usepackage{float}

\definecolor{cvprblue}{rgb}{0.21,0.49,0.74}
\usepackage[pagebackref,breaklinks,colorlinks,citecolor=cvprblue]{hyperref}
\newcommand{\blfootnote}[1]{\begingroup
	\renewcommand\thefootnote{}\footnote{#1}\addtocounter{footnote}{-1}
	\endgroup}

\title{Towards Automatic Power Battery Detection:\\
 New Challenge, Benchmark Dataset and Baseline}

\author{Xiaoqi Zhao$^{1,2\dagger}$, 
	Youwei Pang$^{1,2\dagger}$, 
	Zhenyu Chen$^1$,  
	Qian Yu$^1$, \\
	Lihe Zhang$^{1*}$, 
	Hanqi Liu$^2$, 
	Jiaming Zuo$^{2*}$, 
	Huchuan Lu$^1$\\
	$^1$Dalian University of Technology \quad
	$^2$X3000 Inspection Co., Ltd \quad \\
	{\tt\small
		\{zxq, lartpang, dlutczy, ms.yuqian\}@mail.dlut.edu.cn,
		\{jerry, klaus\}@3000gy.com}\\
	{\tt\small
		\{zhanglihe, lhchuan\}@dlut.edu.cn}
}
	
\begin{document}
\maketitle
\blfootnote{$\dagger$ Equal contribution.}
\blfootnote{$*$ Corresponding author.}
\begin{abstract}
We conduct a comprehensive study on a new task named power battery detection (PBD), which aims to localize the dense cathode and anode plates endpoints from X-ray images to evaluate the quality of power batteries. 
Existing manufacturers usually rely on human eye observation to complete PBD, which makes it difficult to balance the accuracy and efficiency of detection. 
To address this issue and drive more attention into this meaningful task, we first elaborately collect a dataset, called 
X-ray PBD, which has $1,500$ diverse X-ray images selected from thousands of power batteries of $5$ manufacturers, with $7$ different visual interference. 
Then, we propose a novel segmentation-based solution for PBD, termed multi-dimensional collaborative network (MDCNet). With the help of line  and counting predictors, the representation of the point segmentation branch can be improved at both semantic and detail aspects.
Besides, we design an effective distance-adaptive mask generation strategy, which can alleviate the visual challenge caused by the inconsistent distribution density of plates to provide MDCNet with stable supervision.
Without any bells and whistles, our segmentation-based MDCNet consistently outperforms various other corner detection, crowd counting and general/tiny object detection-based solutions, making it a strong baseline that can help facilitate future research in PBD. Finally, we share some potential difficulties and works for future researches. The source code and datasets will be publicly available at \href{https://github.com/Xiaoqi-Zhao-DLUT/X-ray-PBD}{X-ray PBD}.

\end{abstract}
\section{Introduction}
\begin{figure}[t]
\centering
\includegraphics[width=\linewidth]{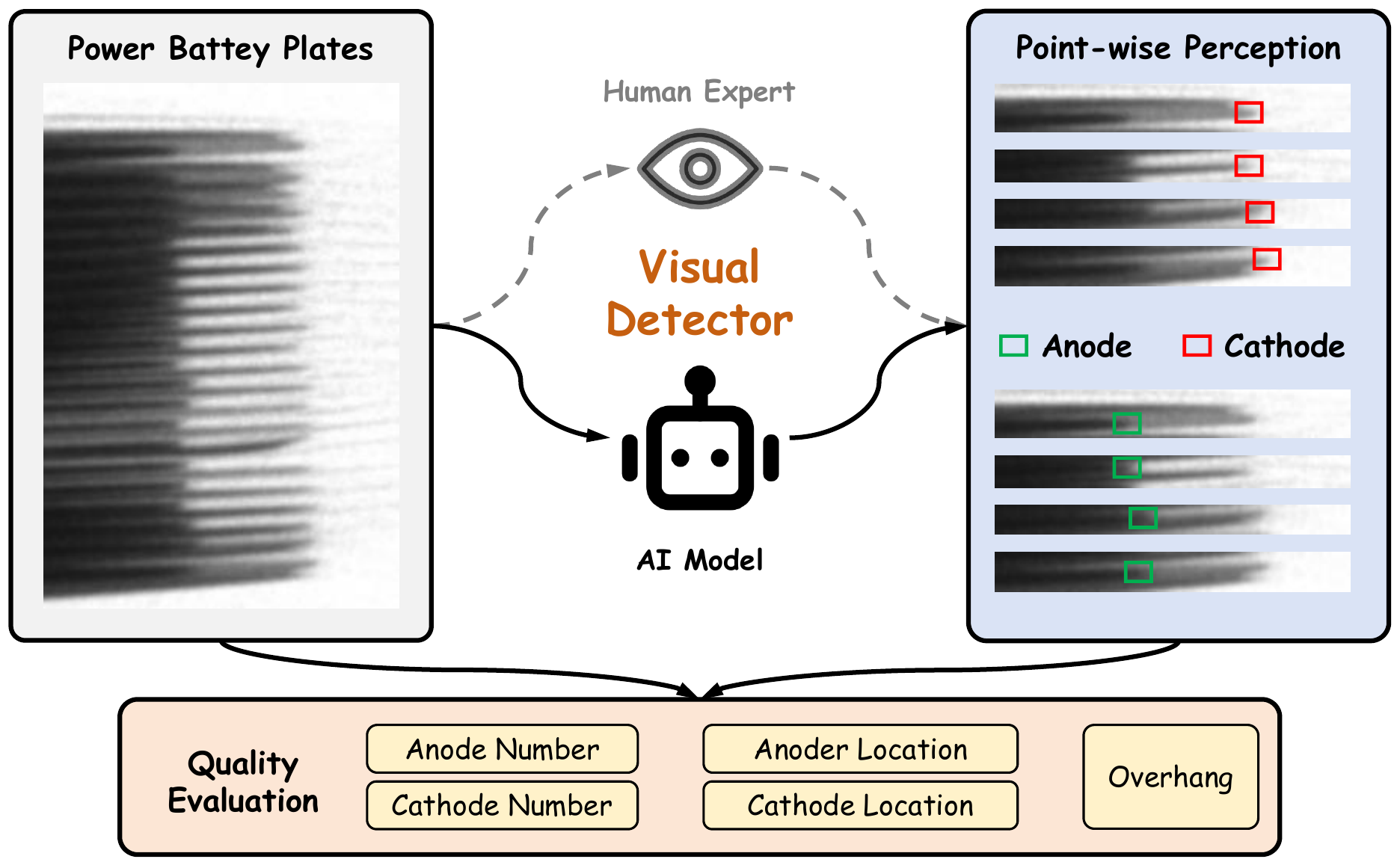}
\caption{Illustration of the power battery detection task. }
\label{fig:task_description}
 \vspace{-5.5mm}
\end{figure}

With the development of power battery technology, new energy vehicles
are receiving more and more attention. 
The power battery is the only source of driving energy for battery electric vehicle (BEV), which directly affects the power performance, endurance and safety of BEV~\cite{BEV_1}. 
To ensure the safety of power battery, the functional evaluation has to be done through  power battery detection (PBD).
As shown in Fig.~\ref{fig:task_description}, the PBD can provide accurate coordinate information for all anode and cathode endpoints. With the help of digital radiography (DR) device, the internal shape of the battery cell can be obtained from the X-ray image, thereby providing humans with the number and overhang \footnote{Number and overhang are two criteria for power battery manufacturers to evaluate battery quality. Number: The number of both anode and cathode plates. Overhang: The average value of the vertical distance between a single anode plate and two adjacent cathode plates.} information to judge whether the current battery cell is OK or NG. 
Most factories evaluate the quality of power batteries through human eyes. However, visual fatigue will be aggravated with long-term work. Although some top battery factories have a large number of re-inspection technicians, this PBD manner is high-cost and low-efficiency.

\begin{figure*}[t]
\centering
\includegraphics[width=\linewidth]{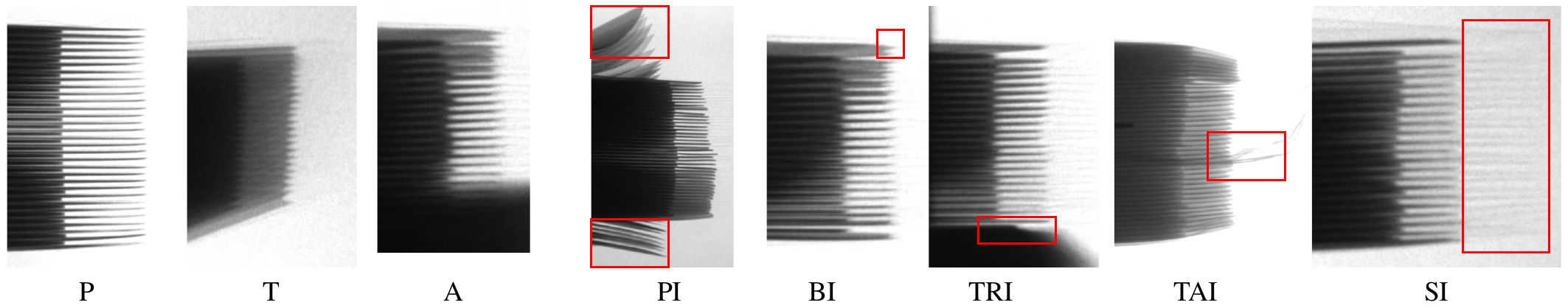}
\caption{Examples of various attributes from our X-ray PBD dataset (best viewed zoomed in). See Tab.~\ref{tab:attribute_description} for details.}
\label{fig:buttary_class}
\end{figure*}

We believe that it is imminent to explore an intelligent PBD model.
To this end,
we build a X-ray PBD dataset via the high-quality industry DR device under thousands of different batteries. It has several characteristics: (1) rich shapes and complex interference (See Fig.~\ref{fig:buttary_class}); (2) vertex coordinates of plates as the annotated ground truth (GT);  (3) widely distributed number of plates; (4) widely distributed overhangs between each pair of anode and cathode plates; (5) various views with long, medium and close shots.  
{The main challenges for computer vision community are: (1) \textit{\textbf{Open vision modeling problem.}} Like emerging tasks such as character stroke extraction~\cite{CCSE}, insubstantial object detection~\cite{IOD}, open vocabulary objects detection and segmentation~\cite{OVD,OVSeg}, how to model a suitable AI-based solution is critical for the PBD task. (2)  \textit{\textbf{Improving the discriminativeness of weak features in deep networks.}} This task aims to accurately locate each independent plate endpoint (1 pixel) from the entire image (millions of pixels), which likes finding a needle in a haystack. In Fig.~\ref{fig:buttary_class}, we can see that the features of plate endpoints are weak and similar with around regions. It is difficult to capture all the endpoints in the face of interference from the plate, bifurcation and separator. Therefore, how to leverage useful cues to  boost weak features is another important challenge. 

In this paper, we propose a multi-dimensional collaborative network (MDCNet) for accurate PBD. \textbf{\textit{First}}, we cast the PBD problem into the segmentation task and build a multi-task framework on multiple dimensions of point, line and number.  Point segmentation predictor produces the position of the endpoint of each plate one by one in the form of a point map. Line segmentation and counting predictors are guided by the point prediction map to separately aggregate low-level and high-level features to predict the trend line and number of plates. Here, the point segmentation branch is the core, and the other two auxiliary tasks utilize spatial and semantic cues to simultaneously fine-tune the point branch.
\textbf{\textit{Next}}, we choose the clear X-ray Image with a standard battery shape as the prompt to generate the prompt filter for the point prediction branch. In this way, our PBD model can have the anti-interference ability and enhance the discrimination of plates. 
\textbf{\textit{Last}}, we calculate the spatial distance of adjacent homopolar plates based on the annotations and use it as the reference radius of the point map to adaptively generate the ground truth for point segmentation. 

Our main contributions can be summarized as follows: 
\begin{itemize}[noitemsep,nolistsep]
\item We propose a new challenging task named power battery detection (PBD) and construct a complex PBD dataset, design an effective baseline, formulate comprehensive metrics, and explore label generation strategies to promote research on the PBD.
\item We formulate the PBD problem into a segmentation task and develop a novel multi-dimensional collaborative network (MDCNet), which leverages the point, line and number cues through the multi-scale feature fusion and prompt filter techniques to achieve accurate PBD under complex and diverse background interference scenes.
\item We compare our segmentation-based method with the other potential solutions, including corner detection, crowd counting and general/tiny object detection. Extensive experiments demonstrate that the proposed MDCNet performs favorably against the state-of-the-art methods under eight metrics. 
\end{itemize}


\begin{table}
	\setlength{\abovecaptionskip}{2pt}
\centering
\resizebox{\linewidth}{!}{
    \setlength\tabcolsep{1pt}
    \renewcommand\arraystretch{1.2}
   \begin{tabu}{cccccl}
\toprule[2pt]
 Attr & &&&&Description\\
 \hline
\textbf{P} &&&&& \textit{Pure Plate.} High-quality  sample without any internal and external interference.\\
\textbf{T} &&&&& \textit{Tilted Plate.} Deformation caused by excessively dense plates.\\
\textbf{A} &&&&& \textit{Aberrant Plate.} 1) Plates visualization is incomplete due to occlusion.\\
&&&&& 2) Plates are not arranged in the order of anode $\rightarrow$ cachode $\rightarrow$ anode. \\
\textbf{PI} &&&&& \textit{Plate Interference.} Other batteries blend into current view.\\
\textbf{BI} &&&&& \textit{Bifurcation Interference.} A single plate produces a bifurcation.\\
\textbf{TRI} &&&&& \textit{Tray Interference.} A battery tray is a device used for holding large battery packs in place. \\
&&&&& Close distance between the side plate and tray may cause visual interference.\\
\textbf{TAI} &&&&& \textit{Tab  Interference.} Battery tabs are the anode and cachode connectors that \\
&&&&& carry the cells' electrical current.  \\
\textbf{SI} &&&&& \textit{Separator Interference.} A battery separator is a type of polymeric membrane \\
&&&&& that is positioned between the anode and cathode. \\

\bottomrule[2pt]
\end{tabu}
	}
 \caption{Attribute descriptions (see examples in Fig.~\ref{fig:buttary_class}).}
\label{tab:attribute_description}
\end{table}

\section{Related Work}
In this section, we introduce multiple possible modeling schemes for the PBD task and analyze related techniques.
\subsection{Corner Detection}
A corner point is usually defined as the intersection of two edges. Corner detection is widely used in motion detection, image matching and video tracking. It can also be called feature point detection. Harris and Stephens~\cite{Harris} propose one of the earliest corner detection algorithms by considering the differential of the corner score with respect to direction directly. Shi–Tomasi~\cite{Shitomasi} corner detector directly computes minimum euclidean distance between corners detected because under certain assumptions, the corners are more stable for tracking. Sub-pixel~\cite{Sub-Pixel} corner detection can obtain real coordinate values to provide more accuracy for performing geometric measurement or calibration. Corner detectors are not usually very robust and often require large redundancies introduced to prevent the effect of individual errors from dominating the recognition task.

\subsection{General/Tiny Object Detection}
Object detection task deals with detecting instances of semantic objects of a certain class (such as humans, buildings, or cars) in digital images and videos. Object detectors usually can be divided into two types:  1) two-stage detectors: Fast R-CNN~\cite{Fast_rcnn}, Faster R-CNN~\cite{Faster_rcnn}, FPN~\cite{FPN}, R-FCN~\cite{RFCN}. 2) one-stage detectors: RetinaNet~\cite{RetinaNet}, YOLOV5~\cite{YOLOv5}, FCOS~\cite{FCOS}, DETR~\cite{DETR}. Although most single-stage methods have fast detection speed and can learn the generalized features of objects, they are usually not as accurate as the anchor-based two-stage detectors, especially in small objects. To this end, some tiny object detectors have been proposed to achieve the balance between accuracy and speed, e.g., MRDet~\cite{MRDet}, QueryDet~\cite{Querydet}, C3Det~\cite{C3Det} and CFINet~\cite{CFINet}. 
\subsection{Dense Object Counting}
Dense object counting task is to infer the number of objects in dense scenarios. Most CNNs-based methods~\cite{DM,CUT_cc,scale_variation_cc1,density_map_cc1,context_information_cc1} predict a density map from a crowd image, where the summation of the density map is the crowd count. 
For the structures, some works are proposed to address scale variation~\cite{scale_variation_cc1,scale_variation_cc2,scale_variation_cc3}, to refine the predicted density map~\cite{density_map_cc1,density_map_cc2,density_map_cc3,IOCFormer}, and to encode context information~\cite{context_information_cc1,context_information_cc2}.
For the loss function, L2 loss is usually used in many models but it is sensitive to the choice of variance in the Gaussian kernel~\cite{Adaptive_CC}. Therefore, Bayesian loss~\cite{BL} is proposed with point-wise supervision. 
\subsection{Image Segmentation}
Image segmentation aims to simplify and/or change the representation of an image into something that is more meaningful and easier to analyze. There are many kinds of image segmentation in real life, such as salient object segmentation, camouflaged object segmentation, glass segmentation and medical image segmentation. Most methods~\cite{GateNet,MINet,MSNet_Polyp,MMFT,MSAPS,HDFNet} tend to adopt U-shape~\cite{Unet,FPN} with the encoder and decoder as the baseline and then combine multi-level features to gradually reconstruct the high-resolution feature maps. To improve the representation ability, some methods~\cite{AMP,BDRAR_Shadow,GateNet,DANet} introduce the attention or gate mechanism and propose different spatial attention and channel-wise attention modules. Besides, objects with size-varying is another challenge in the segmentation filed. This motivates many efforts~\cite{GateNetv2,ASPP,DenseASPP,GateNet,M2SNet,ZoomNet} to investigate multi-scale feature extraction module, such as the ASPP~\cite{ASPP}, Fold-ASPP~\cite{GateNet} and DenseASPP~\cite{DenseASPP}. 

\begin{figure*}[t]
\centering
\includegraphics[width=\linewidth]{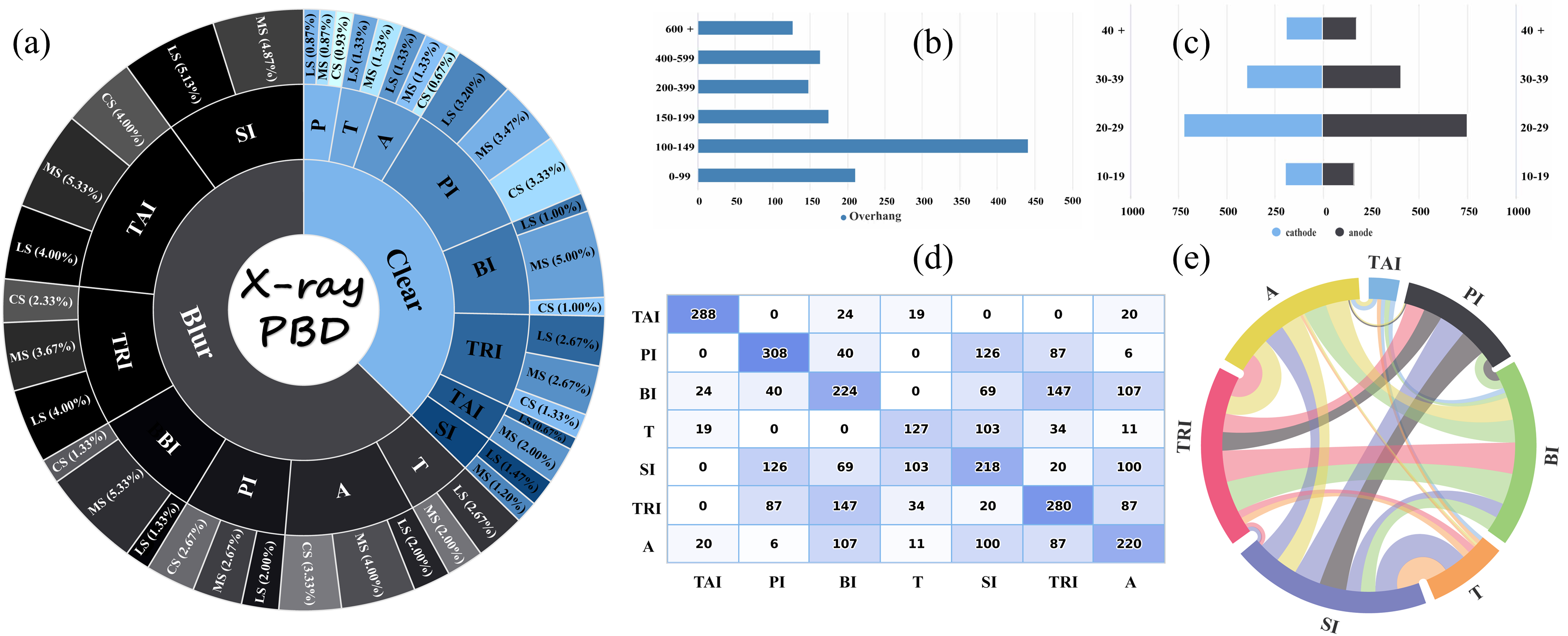}
\caption{Statistics of the X-ray PBD dataset. (a) Taxonomic of interference and shots. (b) Overhang distributions. (c) Number distributions. (d) Co-occurrence  distribution of attributes. (e) Multi-dependencies among these attributes.}
\label{fig:datasets_characteristic}
\end{figure*}

\section{X-ray PBD Dataset}
\subsection{Image Collection and Annotation}
The proposed X-ray PBD dataset is collected from thousands of power batteries cells of $5$ manufacturers captured by the same DR device, covering close, medium, and long shots. It is divided into $2$ plate classes with $8$ attributes according to appearance and interference types. We first deliver the initially collected $4,000 +$ X-ray battery images to six professional battery factory technicians for quality screening. Through careful observation, duplicate, blank, and severely defective (i.e., invalid) images are eliminated and we obtain $\sim 2,000$ battery images. Then, six technicians use the labeling software to annotate $2,000$ images, respectively. The annotation information include the endpoint coordinates of all anode and cathode plates, and the numbers of plates. To avoid human judgment errors, six technicians jointly evaluate all six sets of marked data and choose the examples with consistent marked information. Finally, we obtain $1,500$ high-quality power battery images. 
\subsection{Dataset Features, Statistics and Challenges}
$\bullet$~\textit{Categories.}
As illustrated in Fig.~\ref{fig:datasets_characteristic} (a), the collected images have $2$ categories and $8$ attributes and their proportions in each shot type are summarized in detail. Different battery types, manufacturing and packaging processes produce different thicknesses of batteries. Even if the DR device is calibrated, it still has different degrees of penetration for different thicknesses, thereby resulting in  ``clear'' and ``blur'' X-ray images.  Attributes descriptions and visualization are provided in Tab.~\ref{tab:attribute_description} and Fig.~\ref{fig:buttary_class}.

\noindent$\bullet$~\textit{Number and Overhang.}
Generally speaking, different types of batteries have different numbers of plates. Even in the same type of product, overhang is not exactly the same. As shown in Fig.~\ref{fig:datasets_characteristic} (b, c), the X-ray PBD dataset covers a large range of overhang and number distributions.

\noindent$\bullet$~\textit{Attributes.}
Fig.~\ref{fig:datasets_characteristic} (d, e) demonstrate co-occurrence distributions and dependencies among these attributes, respectively. 
We can see that more than four attribute dependencies appear in all samples, which illustrates the diversity and complexity of this dataset.

\noindent$\bullet$~\textit{Dataset splits.}
To provide a large amount of training
data for deep learning models, X-ray PBD is split into $900$
images for training and $600$ for testing, randomly selected from each sub-class with different attributes. And, we further divide the test set into regular (239 images), difficult (187 images) and tough (174 images) subsets according to the interference degree caused by different attributes.

\noindent$\bullet$~\textit{Challenges.}
To realize effective power battery detection, it faces multiple challenges caused by battery characteristics, photography restrictions, internal and external interference. First, there exist three types of shots in the X-ray PBD dataset. A well-performed PBD model needs to perceive the size-varying battery regions
sensitively. Second, various interference and their dependencies  make the discriminative information of the plates inconspicuous and easily overwhelmed by disturbances, which requires the model to have strong anti-interference ability. Third, both intra-sample and inter-sample plates have different overhangs, it brings the PBD model with strict requirements on both semantic discrimination and spatial localization. Fourth, the error tolerance rate is extremely low. In the real-life industrial production, all plates of a single battery cell must be detected with 100\% accuracy, otherwise the automatic inspection is meaningless.
\subsection{Evaluation Metrics}
According to the number and overhang criteria adopted by the manufacturers, we design eight complementary metrics to quantitatively evaluate the performance of algorithms. Specifically, they are number mean absolute error (AN-MAE, CN-MAE), number accuracy (AN-ACC, CN-ACC, PN-ACC) and position mean absolute error (AL-MAE, CL-MAE and OH-MAE) for the anode level, cathode level, and pair level, respectively. Note that,  
for an image, only when its number of AN or CN is predicted with 100\% accuracy, we can further evaluate the corresponding position MAE. More details are provided in appendix.

\begin{figure*}[t]
\centering
\includegraphics[width=\linewidth]{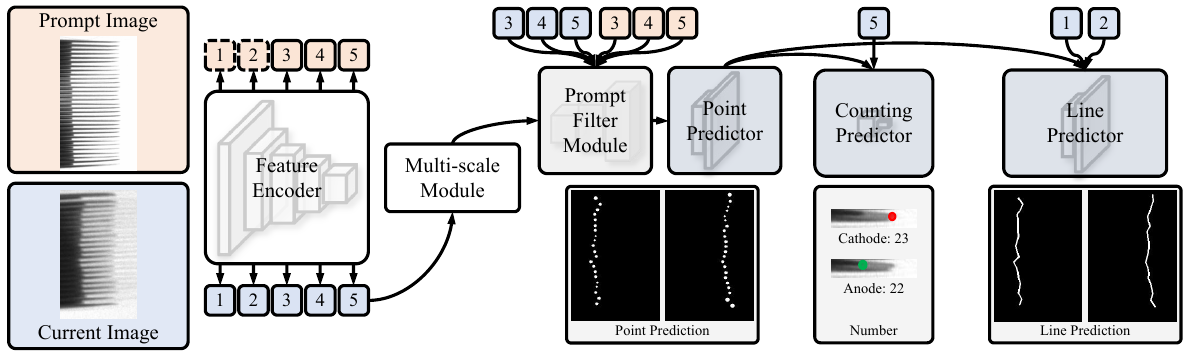}
\caption{Overview of our MDCNet. It contains a shared encoder to extract different level features for the prompt and current images, respectively. Multi-scale module is only embedded in the high-level features. Prompt filter module are used to combine the prompt and current features to generate a series of filtered features. Point predictor include five decoder layers to produce point segmentation maps. Counting and line predictors are guided by the point prediction and fusing high-level and low-level features, respectively. }
\label{fig:Network_1}
\end{figure*}

\begin{figure*}[t]
\centering
\includegraphics[width=\linewidth]{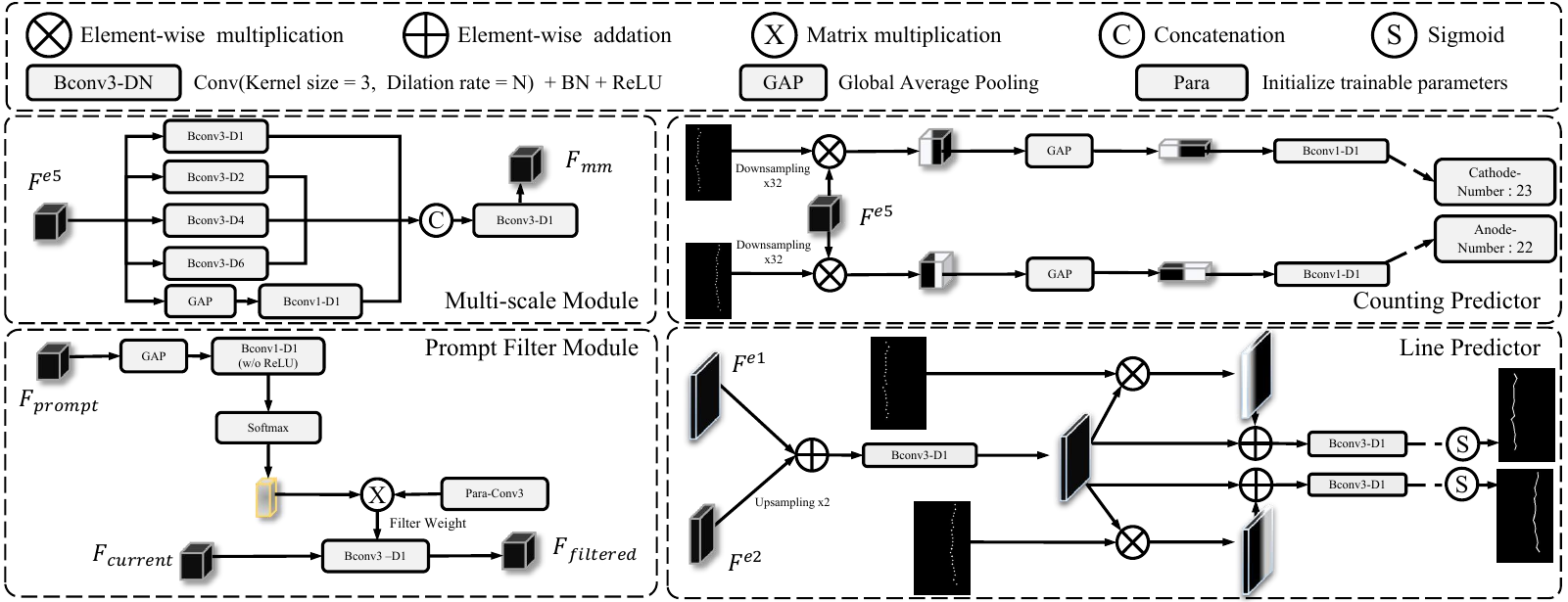}
\caption{Detailed illustration of each component.}
\label{fig:components}
\end{figure*}

\section{Proposed Framework}
\subsection{Multi-dimensional Collaborative Network}
\noindent\textbf{Motivation}. 
Imitating the coarse-to-fine human visual perception system~\cite{coarse-to-fine1,coarse-to-fine2}, the PBD model first perceives the battery area, and then gradually identifies each plate. Once the endpoint of each plate is accurately segmented, the position coordinates and overall quantity information can further be obtained. General U-shape~\cite{FPN,Unet} structures are adopted as the point branch to generate initial point prediction.  Inspired by multi-task learning~\cite{Hard_parameter_sharing,soft_parameter_sharing,MMFT} and characteristics of power battery, we extra design line segmentation and counting tasks to constrain and tune the point segmentation branch, helping it to complete the prediction from coarse to fine. The predictions of three dimensions of point, line and number cooperate with each other, dubbed as multi-dimensional collaborative network (MDCNet).

\noindent\textbf{Overview}. 
As shown in Fig.~\ref{fig:Network_1}, the MDCNet has the encoder-decoder structure  with the ResNet-50~\cite{Resnet} backbone. 
The encoder first extracts five levels of features, meanwhile, prompt and current pipelines share the same weights. 
Multi-scale module and the prompt filter module are embedded into the point segmentation branch. 
The former integrates features with receptive fields of different scales for improving the perception of multi-scale objects.
The latter utilizes the prompt features to filter interference information from the current features and generate pure plate appearance features. 
Point predictor outputs the point segmentation map after passing through five decoder blocks.
Line and counting predictors assist point prediction at low-level details and semantic level, respectively.
\subsection{Components}
As shown in Fig.~\ref{fig:components}, it illustrates each  component in the MDCNet, including the multi-scale module, prompt filter module, counting predictor and line predictor. 

\noindent$\bullet$~\textit{Multi-scale module.}
The single-scale features can not capture the multi-scale context for batteries with different shapes under different shots.  
We apply the multi-scale module to the high-level features, which have rich semantic cues for localizing objects. The multi-scale module includes four convolution branches with dilation rates of $[1, 2, 4, 6]$ and a global average pooling branch. The features with different receptive fields are fused by cross-channel concatenation.

\noindent$\bullet$~\textit{Prompt filter module.} 
In order to weaken external interference such as trays, bifurcation, tabs, etc., we embed the prompt filter module (PFM) in the point segmentation branch. We randomly choose a \textit{P} plate sample (see Fig~\ref{fig:buttary_class}) as the prompt image. 
The 3$^{rd}$-5$^{th}$ layer features of the prompt image will be separately generated the filters to edit the current image features, which progressively fuses the different levels of features through five convolution layers. 
We take both the 5$^{th}$ prompt and current image features as an example to illustrate the process of prompt filtering. 
As shown in Fig.~\ref{fig:components}, we first conduct global average pooling and convolution for $F_{prompt}$. Then, the softmax function distributes  $F_{prompt}$ with the channel-wise soft attention. Next, we initialize the  $3\times3$ convolution with trainable parameters and multiply the soft attention to generate the aggregated parameters as the weights. Finally, we obtain the enhanced feature $F_{filtered}$, which is  passed to the point predictor.

\noindent$\bullet$~\textit{Counting predictor.}
The overall counting predictor is constructed to solve a regression problem. The inputs of the counting predictor are the high-level features ($F_{current}^{e5}$)  and the predicted 
 point maps ($M^{anode}_{p}$, $M^{cathode}_{p}$) of the anode and cathode plates from the point predictor. To narrow the search range of the counting objects, we utilize two predicted point maps as spatial attention to guide $F_{current}^{e5}$. The number of anode and cathode plates are computed as:
\begin{equation}\label{equ:1}
\centering
\small
\left\{\begin{matrix}
    N^{anode} = ReLU(Conv(GAP(DS(F_{current}^{e5}) \otimes M^{anode}_{p})))\\
 N^{cathode} = ReLU(Conv(GAP(DS(F_{current}^{e5}) \otimes M^{cathode}_{p}))),\\
    \end{matrix}\right.
\end{equation} 
where DS(·) is the downsampling operation, $\otimes$ is the element-wise multiplication, GAP(·) is the global average pooling and Conv(·) refers to the convolution layer with the one channel output. 
As an auxiliary task for the point segmentation, counting task can improve the query ability of high-level features on the number of plates at the global level, thereby enhancing the feature representation of the point branch.

\noindent$\bullet$~\textit{Line predictor.}
Generally speaking, detail information of contours often exists in the low-level features. Therefore, we build the line predictor with both low-level features ($F_{current}^{e1}$, $F_{current}^{e2}$) and the predicted 
 point maps ($M^{anode}_{p}$, $M^{cathode}_{p}$) as inputs. 
We first aggregate the $F_{current}^{e1}$, $F_{current}^{e2}$ to generate the $F_{current}^{e1,2}$:
\begin{equation}\label{equ:2}
\centering
\small
\begin{matrix}
    F_{current}^{e1,2} = Conv(F_{current}^{e1} + US(F_{current}^{e2})), 
\end{matrix}
\end{equation} 
where US(·) is the upsampling operation. Next, $F_{current}^{e1,2}$ is separately computed with  $M^{anode}_{p}$ and $M^{cathode}_{p}$
in the form of residual~\cite{Resnet} and predict line segmentation maps:
\begin{equation}\label{equ:3}
\centering
\small
\left\{\begin{matrix}
    L^{anode} = S(Conv(M^{anode}_{p} \otimes F_{current}^{e1,2}) + F_{current}^{e1,2})\\
  L^{cathode} = S(Conv(M^{cathode}_{p} \otimes F_{current}^{e1,2}) + F_{current}^{e1,2}),\\
    \end{matrix}\right.
\end{equation} 
where S(·) is the element-wise sigmoid function. 
As another auxiliary task of point segmentation, line segmentation can provide continuous segmentation cues to compensate and modify some plates that are not accurately predicted by point branch.

\begin{figure}[t]
\centering
\includegraphics[width=1.0\linewidth]{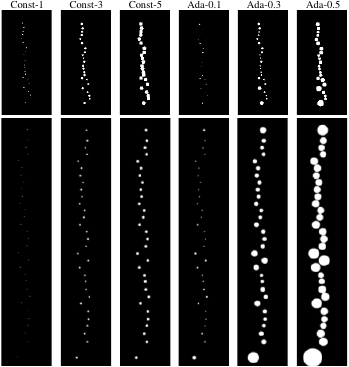}
\caption{Visualization of the ground-truth point masks under different strategies.}
\label{fig:point_mask_strategy}
\vspace{-5mm}
\end{figure}

\begin{table*}[!t]
\centering
	\setlength{\abovecaptionskip}{2pt}
	
   \resizebox{\linewidth}{!}{
    \setlength\tabcolsep{2pt}
    \renewcommand\arraystretch{1.2}
    \begin{tabular}{|c|r||ccc|ccc|ccccc|c|}
     \hline
      &  & \multicolumn{3}{c|}{\textbf{{Corner Detection}}}& \multicolumn{3}{c|}{\textbf{{Crowd Counting }}}& \multicolumn{5}{c|}{\textbf{{General/Tiny Object Detection
 }}} & \textbf{Segmentation}\\
     Dataset&Metrics  & Harris~\cite{Harris} & Shi-tomasi~\cite{Shitomasi}
&Sub-pixel~\cite{Sub-Pixel}& BL~\cite{BL} & CUT~\cite{CUT_cc} & IOCFormer~\cite{IOCFormer} & RetinaNet~\cite{RetinaNet} & Faster RCNN~\cite{Faster_rcnn}
 &YOLOv5~\cite{YOLOv5}&C3Det~\cite{C3Det}&CFINet~\cite{CFINet} &MDCNet (Ours)
     \\
   				\hline
				\hline
     \multirow{8}{*}{Regular} 
     &AN-MAE$\downarrow$
     &125.4128
     &81.2936&116.4404&	0.3211
&0.3028
&0.8349
 &	0.6330 &	1.2294 &	0.3761 	&	3.1835 &	0.5505 &	\color{reda}\textbf{0.0642}
 \\
   &CN-MAE$\downarrow$
     &3437.5780
&156.9083&121.7615&0.1468
&0.8349
&0.3853
&	0.3486&	0.7798 &	0.5413	&	0.5138 &	0.2477 &	\color{reda}\textbf{0.0183}
 \\
   &AN-ACC$\uparrow$
     &0.0000
&0.0000
&0.0000
     &0.6881
&0.6972
&0.3211
&0.6330
&0.4220
&0.6881
&0.6330
&0.7339
&\color{reda}\color{reda}\textbf{0.9633}
 \\
  &CN-ACC$\uparrow$
    & 0.0000
&0.0000
&0.0000
&0.8532
&0.1743
&0.6514
&0.7615
&0.6972
&0.6972
&0.8532
&0.8991
&\color{reda}\textbf{0.9908}
 \\
  &PN-ACC$\uparrow$
    &0.0000
&0.0000
&0.0000
&0.5872
&0.1193
&0.2294
&0.4587
&0.3761
&0.4954
&0.6330
&0.6881
&\color{reda}\textbf{0.9541}
 \\
  &AL-MAE$\downarrow$
    &--
&--
&--
&--
&--
&--
&6.6782
&16.4639
&5.1243
&4.9344
&4.0220
&\color{reda}\textbf{2.3365}
\\
 &CL-MAE$\downarrow$
    &--
&--
&--
&--
&--
&--
&6.0135
&14.8421
&4.6982
&3.6087
&3.8072
&\color{reda}\textbf{1.8411}

 \\
&OH-MAE$\downarrow$
    &--
&--
&--
&--
&--
&--
&3.2207
&6.4742
&5.1286
&4.7642
&3.9496
&\color{reda}\textbf{2.0422}
 \\
\hline
    \multirow{8}{*}{Difficult} 
     &AN-MAE$\downarrow$
     &122.1703
&73.5899
&106.4858
&1.3211
&2.4216
&0.9117
&1.1546
&1.5047
&1.5047
&2.2650
&1.0883
&\color{reda}\textbf{0.4259}
 \\
   &CN-MAE$\downarrow$
     &3208.6183
&146.3344
&113.3249
&1.1468
&3.3215
&0.5079
&0.7476
&1.1546
&1.7192
&1.1703
&0.8139
&\color{reda}\textbf{0.2050}

 \\
   &AN-ACC$\uparrow$
     &0.0000
&0.0000
&0.0000
&0.7382
&0.5584
&0.3659
&0.3975
&0.4038
&0.4416
&0.5616
&0.5741
&\color{reda}\textbf{0.7855}
 \\
  &CN-ACC$\uparrow$
    &0.0000
 &0.0000
 &0.0032
 &0.8170
&0.4479
&0.5552
 &0.5804
 &0.5584
 &0.4069
 &0.7035
 &0.8991
 &\color{reda}\textbf{0.9306}

 \\
  &PN-ACC$\uparrow$
    &0.0000
&0.0000
&0.0000
&0.6278
&0.2965
&0.2461
&0.2587
&0.3091
&0.2177
&0.5016
&0.5426
&\color{reda}\textbf{0.7603}

 \\
  &AL-MAE$\downarrow$
    &--
&--
&--
&--
&--
&--
&6.9021
&28.0690
&4.1363
&4.9417
&4.9602
&\color{reda}\textbf{2.4397}

\\
 &CL-MAE$\downarrow$
    &--
 &--
 &44.5243
 &--
 &--
 &--
 &6.4768
 &23.7668
 &4.0355
 &4.7920
 &4.9878
 &\color{reda}\textbf{2.0978}

 \\
&OH-MAE$\downarrow$
    &--
&--
&--
&--
&--
&--
&3.4025
&6.9621
&3.1810
&3.7909
&3.9769
&\color{reda}\textbf{2.1092}
 \\
\hline
    \multirow{8}{*}{Tough} 
     &AN-MAE$\downarrow$
     &59.7989
&16.8621
&20.0287
&2.5989
&2.8212
&2.3036
&4.1552
&5.0057
&6.5632
&3.7356
&3.2989
&\color{reda}\textbf{2.0920}

 \\
   &CN-MAE$\downarrow$
     &1132.0632
&22.0862
&18.9080
&2.1494
&3.2275
&2.1044
&3.4253
&4.2011
&6.1437
&2.5747
&3.1494
&\color{reda}\textbf{1.8966}

 \\
   &AN-ACC$\uparrow$
     &0.0000
&0.0000
&0.0172
&0.3448
&0.3103
&0.5057
&0.2931
&0.4425
&0.3448
&0.3621
&0.3793
&\color{reda}\textbf{0.5632}
 \\
  &CN-ACC$\uparrow$
    &0.0000
&0.0115
&0.0172
&0.4023
&0.2872
&0.5230
&0.5230
&0.6494
&0.2126
&0.4195
&0.4310
&\color{reda}\textbf{0.6839}

 \\
  &PN-ACC$\uparrow$
    &0.0000
&0.0000
&0.0000
&0.3161
&0.2586
&0.3218
&0.1954
&0.3678
&0.0690
&0.3046
&0.3276
&\color{reda}\textbf{0.5115}
 \\
  &AL-MAE$\downarrow$
    &--
&--
&85.5647
&--
&--
&--
&4.9547
&9.9379
&4.0577
&4.0700
&4.9448
&\color{reda}\textbf{2.0004}
\\
 &CL-MAE$\downarrow$
    &--
&53.0062
&51.1617
&--
&--
&--
&4.9750
&7.2092
&4.4330
&4.6340
&4.6617
&\color{reda}\textbf{1.4654}
 \\
&OH-MAE$\downarrow$
    &--
 &--
 &--
 &--
 &--
 &--
 &3.8719
 &1.9481
 &4.9554
 &3.6690
 &3.6988
 &\color{reda}\textbf{1.6291}
 \\
\hline
    \end{tabular}
	}
	\setlength{\abovecaptionskip}{2pt}
 \caption{Quantitative comparison of different methods. $\uparrow$ and $ \downarrow$ indicate that the larger scores and the smaller ones  are better, respectively. The best scores are highlighted in {\color{reda} \textbf{red}}. ``—'' represents that the results are not available because these methods can not provide coordinate information or their prediction accuracy of the number of plates is zero. }
	\label{tab:comparison}
\end{table*}

\begin{figure*}[!t]
\centering
\includegraphics[width=\linewidth]{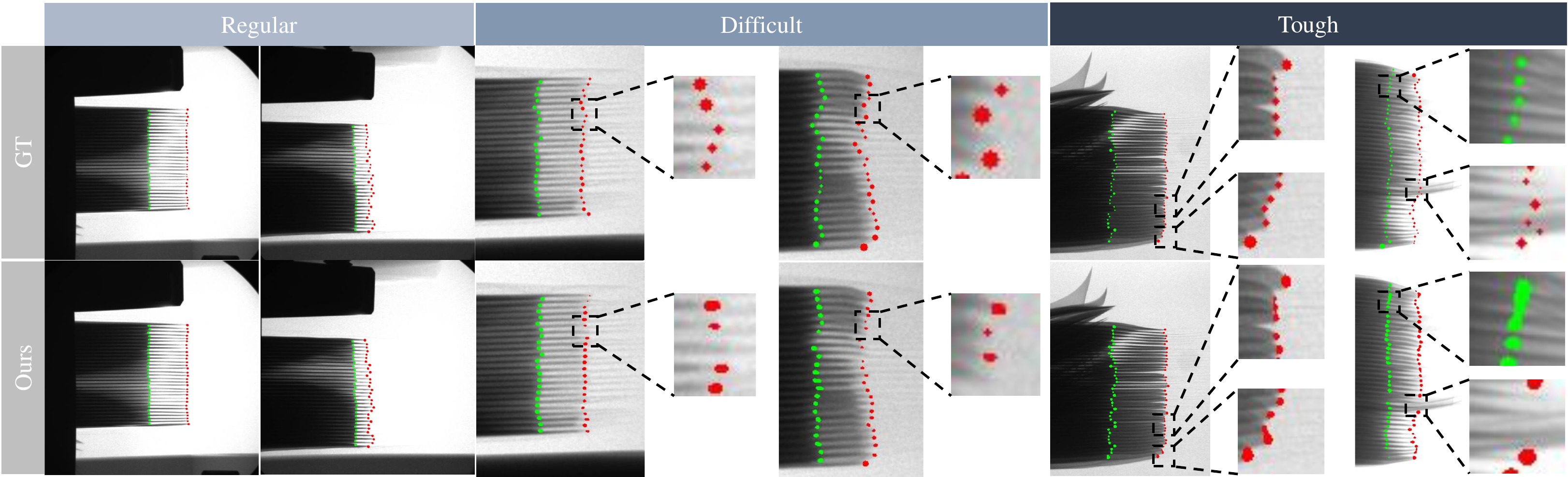}
\caption{Qualitative results on the regular, difficult and tough examples with different shots and attributes. Best viewed zoomed in.}
\label{fig:visual_results}
\end{figure*}

\begin{figure*}[!t]
\centering
\includegraphics[width=\linewidth]{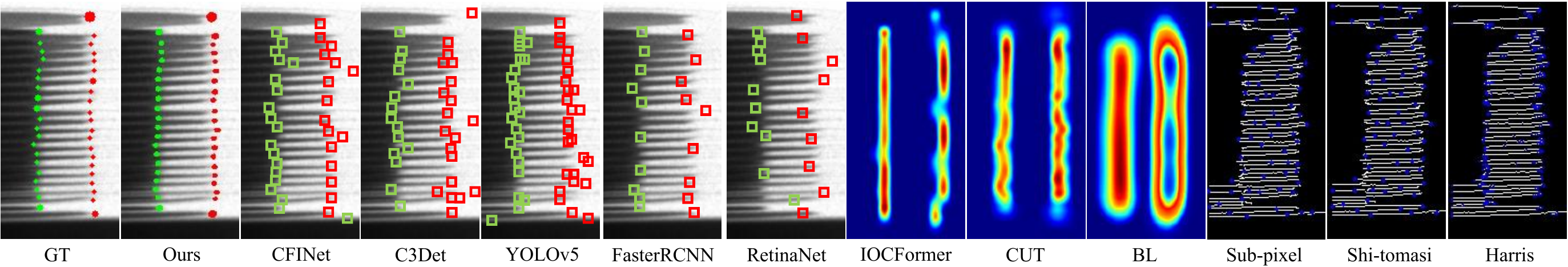}
\caption{Visual comparison with other general/tiny object detection-based~\cite{CFINet,C3Det,YOLOv5,Faster_rcnn,RetinaNet}, counting-based~\cite{BL,CUT_cc,IOCFormer}, corner detection-based~\cite{Sub-Pixel,Shitomasi,Harris} solutions. We directly visualize the predicted results (Ours: Segmentation map, General/Tiny object detection methods: Bounding box, Counting methods: Density map, Corner detection methods: Corner map) without any post-processing operations.}
\label{fig:visual_comparison}
\vspace{-3mm}
\end{figure*}

\subsection{Label Generation and Supervision}\label{sec:generation_mask}
For point segmentation, a direct strategy of generating the ground truth of point mask is to use circular regions of the same radius for each plate. As shown in the $1^{st}$ - $3^{rd}$ columns of Fig.~\ref{fig:point_mask_strategy}, the point masks are yielded under the radius of $1$, $3$, and $5$ pixels, respectively. It can be seen that the fixed radius often produces   very dense or sparse point masks for different types of batteries, which increases the difficulty of point segmentation. To this end, we design a novel distance-adaptive label generation strategy. It computes the diameter of each point according to the distance of adjacent plates. As shown in the $4^{th}$ - $6^{th}$ columns of Fig.~\ref{fig:point_mask_strategy}, the point masks are generated under the diameter of $0.1x$, $0.3x$, and $0.5x$ distance, respectively. The labels of counting and line segmentation can be separately obtained by the connected component analysis and connecting each endpoint in the point mask in sequence.

For the point and line segmentation predictors, we use the weighted IoU loss and binary cross entropy (BCE) loss, which have been widely adopted in segmentation task. We use the same definitions as in~\cite{SPNet,F3Net,PraNet,MSNet_Polyp}. For the counting predictor, we adopt the common L1 loss.

\section{Experiments}
\subsection{Implementation Details}
Our model is implemented based on the Pytorch and trained on one Tesla V100 GPU for $50$ epochs with mini-batch size $12$. The input resolutions of  X-ray images are resized to $352\times352$.  To keep the spatial distribution pattern of the plates, we only adopt the random flipping image augmentation. We use the Adam optimizer~\cite{Adam}.  The initial learning rate  is  set  to $0.0001$. We adopt the ``step'' learning rate decay policy, and set decay size as $30$ and decay rate as $0.9$. Our code and dataset will be made publicly available soon.

\subsection{Comparison with Different Solutions}
We conduct thorough comparisons with corner detection, crowd counting, general/tiny object detection methods. All these models have been retrained on our X-ray PBD dataset based on their publicly released codes. Tab.~\ref{tab:comparison} shows performance comparisons in terms of eight metrics. Our method outperforms all of the competitors and obtains the best performance in terms of all metrics on the regular, difficult and tough splits. It is worth noting that the MDCNet achieves dominant performance in terms of the PN-ACC compared to the second best method CFINet~\cite{CFINet} on the regular ($0.9541$ vs $0.6881$), difficult ($0.7603$ vs $0.5426$) and tough ($0.5115$ vs $0.3276$) subsets. 
Fig.~\ref{fig:visual_results} illustrates visual results of the proposed algorithm on the challenge X-ray images with different complex interference. For regular examples, our model accurately captures the body of each battery, and completes fine semantic segmentation for all the endpoints of anode and cathode plates. For difficult examples, we can overcome most of the separator interference but still lose the sporadic plate that blends well with the separator. For tough examples, the plates are very dense and the roots overlap, which will result in multiple adhesions in the predicted point map.
Fig.~\ref{fig:visual_comparison} illustrates the visual comparison of different solutions with our segmentation-based MDCNet. It can be seen that general/tiny object detection methods often suffer from missed and repeated detection failures due to diverse interference of the bifurcation, tray and separator. For counting methods, the density maps provide rough and overall location of the plates but can not explicitly show each plate one by one. 
For corner detection methods, the corner maps introduce a lot of redundancy and can not distinguish plate endpoints from general intersection points even if we have performed edge extraction to pre-filter a large amount of background irrelevant to plate endpoints.

\begin{table}
\centering
	\scriptsize
	\setlength{\abovecaptionskip}{2pt}
   \resizebox{\linewidth}{!}{
    \setlength\tabcolsep{1pt}
    \renewcommand\arraystretch{1.2}
    \begin{tabular}{|c||c|c|c|c|c|c|c|c|}
\hline
  Method & AN- & CN- & AN- & CN- & PN- & AL- & CL- & OH-      \\
    & MAE$\downarrow$ & MAE$\downarrow$ & ACC$\uparrow$ & ACC$\uparrow$ & ACC$\uparrow$ & MAE$\downarrow$ & MAE$\downarrow$ & MAE$\downarrow$      \\
  \hline \hline      
Baseline
&1.5424
&1.2732
&0.6697
&0.7064
&0.6239
&3.5317
&2.8830
&3.1929
 \\
 + PFM
&0.8211
&0.4324
&0.8257
&0.8440
&0.7706
&3.3298
&2.6425
&3.0231
\\
 + CP
&0.3892
&0.1782
&0.9174
&0.9266
&0.8716
&3.2343
&2.5322
&2.9371
\\
 \rowcolor{mygray}
 + LP
&0.0642
&0.0183
&0.9633
&0.9908
&0.9541
&2.3365
&1.8411
&2.0422
\\
  \hline
\end{tabular}

	}
	\setlength{\abovecaptionskip}{2pt}
 \caption{Ablation experiments of each component in the MDCNet.}
	\label{tab:ablation_study_1}
\end{table}

\begin{table}
\centering
	\scriptsize
	\setlength{\abovecaptionskip}{2pt}

   \resizebox{\linewidth}{!}{
    \setlength\tabcolsep{1pt}
    \renewcommand\arraystretch{1.2}
    \begin{tabular}{|c||c|c|c|c|c|c|c|c|}
  \hline       
  
  Method & AN- & CN- & AN- & CN- & PN- & AL- & CL- & OH-      \\               & MAE$\downarrow$ & MAE$\downarrow$ & ACC$\uparrow$ & ACC$\uparrow$ & ACC$\uparrow$ & MAE$\downarrow$ & MAE$\downarrow$ & MAE$\downarrow$      \\
  \hline \hline      
Const-1
&3.3234
&2.0231
&0.4954
&0.5505
&0.4587
&2.1146
&1.7322
&2.0018

              \\
Const-3
&1.7312
&1.1254
&0.7523
&0.8257
&0.7523
&2.4215
&1.8743
&2.1797

 \\
Const-5
&1.7622
&1.0817
&0.7523
&0.8073
&0.7339
&2.4347
&1.8922
&2.1922

\\
Ada-0.1
&2.0236
&1.4576
&0.6330
&0.6881
&0.6055
&2.4617
&1.9020
&2.1178

\\
 \rowcolor{mygray}
Ada-0.3
&0.0642
&0.0183
&0.9633
&0.9908
&0.9541
&2.3365
&1.8411
&2.0422
\\
Ada-0.5
&0.3452
&0.2971
&0.9174
&0.9450
&0.8991
&2.5301
&1.9420
&2.1201

\\
  \hline
\end{tabular}

	}
	\setlength{\abovecaptionskip}{2pt}
 	\caption{Quantitative comparison of different label generation strategies and settings for the point segmentation branch.}
	\label{tab:ablation_study_2}
 \vspace{-5mm}
\end{table}

\subsection{Ablation Study}
\textbf{Multi-dimensional Collaborative Network.} 
We analyze the contribution of each component to the overall network in Tab.~\ref{tab:ablation_study_1}, which shows the results of ablation study on the regular split. The results on the other two splits can be found in appendix. 
First, our baseline only has the point segmentation branch, which is installed on the FPN~\cite{FPN} with ResNet-50 backbone. We can see that this baseline has been able to suppress most competitors expect the CFNet and C3Det. Hence, the performance gain in subsequent experiments is more convincing. 
Next, we embed the prompt filter module (PFM) into the point segmentation branch. It can be seen that the performance is significantly improved with the gain of $\sim$ 15\% in terms of the AN-ACC, CN-ACC and PN-ACC. 
Finally, we add the counting predictor (CP) and line predictor (LP) to further refine the point segmentation at the high-level and low-level features. The former improves the number accuracy and the latter helps modify the failure positioning. Thus, multi-dimensional features complement each other.

\noindent\textbf{Label Generation Strategies.} 
For point segmentation, we conduct a series of experiments about label generation strategies as discussed in Sec.~\ref{sec:generation_mask}.
As shown in Tab.~\ref{tab:ablation_study_2}, 
the model trained under the Ada-0.3 performs best in terms of number accuracy and localization error among all the six settings. The distance-adaptive  strategy consistently outperforms the constant strategy.

\noindent\textbf{Prompt Filter Module.} 
In Tab.~\ref{tab:ablation_study_4}, we list five groups of results using different \textit{P} attribute images as the anchor of prompt filter module. It can be seen that PFM shows strong stability and generalization without dependence on a specific prompt.

\begin{table}
\centering
	\scriptsize
	\setlength{\abovecaptionskip}{2pt}

   \resizebox{\linewidth}{!}{
    \setlength\tabcolsep{1pt}
    \renewcommand\arraystretch{1.2}
    \begin{tabular}{|c||c|c|c|c|c|c|c|c|}
  \hline
  No. & AN- & CN- & AN- & CN- & PN- & AL- & CL- & OH-      \\               & MAE$\downarrow$ & MAE$\downarrow$ & ACC$\uparrow$ & ACC$\uparrow$ & ACC$\uparrow$ & MAE$\downarrow$ & MAE$\downarrow$ & MAE$\downarrow$      \\
  \hline \hline      
1
&0.0642
&0.0183
&0.9633
&0.9908
&0.9541
&2.3365
&1.8411
&2.0422

              \\
2
&0.0642
&0.0183
&0.9633
&0.9908
&0.9541
&2.3365
&1.8411
&2.0422

 \\
3
&0.0642
&0.0183
&0.9633
&0.9908
&0.9541
&2.3365
&1.8411
&2.0422

\\
4
&0.0642
&0.0183
&0.9633
&0.9908
&0.9541
&2.3365
&1.8411
&2.0422

\\
5
&0.0642
&0.0183
&0.9633
&0.9908
&0.9541
&2.3365
&1.8411
&2.0422

\\
  \hline
\end{tabular}

	}
	\setlength{\abovecaptionskip}{2pt}
 	\caption{Evaluation of the PFM with different prompt inputs.}
	\label{tab:ablation_study_4}
 \vspace{-5mm}
\end{table}

\section{Future Works}
 Because the performance on difficult and tough splits is bad, there remains a long way to go to solve the PBD task. Some potential research are of benefit to the development of the PBD community. {{First}},  
 how to better model PBD is still an open problem. {{Second}}, the annotation of PBD relies on experienced workers and the amount of data is scarce. Semi/self-supervised and few-shot learning techniques may be necessary. 
 {{Third}}, image enhancement techniques such as image super-resolution, restoration and deblurring can be jointly modeled in order to clean the battery plates. 
 {{Last}}, the generality and specificity among multiple scenarios and interference are worth exploring in depth in order to design a robust unified model.

\section{Conclusion}
In this work, 
we present a new task named power battery detection (PBD), which is a crucial course in the new energy industry field.  
We build a complex X-ray PBD dataset, formulate evaluation metrics and propose a segmentation-based solution
(MDCNet). Compared with other solutions based on corner detection, crowd counting and general/tiny object detection, 
MDCNet achieves a dominant performance with over 95\% detection accuracy on regular samples. In the future, more effort will be devoted to handling hard samples. We also plan to extend the PBD dataset to a 3D form with the help of CT device, which can provide richer internal slices information. 

\noindent\textbf{Acknowledgements.}
This work was supported by the National Natural Science Foundation of China under Grant 62276046 and by Dalian Science and Technology Innovation Foundation under Grant 2023JJ12GX015.

\appendix
\section{Appendix}

\subsection{Evaluation Metrics}
There are eight metrics used in this work, as follows:
\begin{equation}\label{equ:1}
\centering
\small
MAE_{num}^{anode}=\frac{1}{N} \sum_{i=1}^{N}\left|n_{i}^{anode}-\widehat{n}_{i}^{anode}\right|,
\end{equation} 
\begin{equation}\label{equ:2}
\centering
\small
MAE_{num}^{cathode}=\frac{1}{N} \sum_{i=1}^{N}\left|n_{i}^{cathode}-\widehat{n}_{i}^{cathode}\right|,
\end{equation} 
\begin{equation}\label{equ:3}
\centering
\small
Acc_{num}^{anode}=\frac{1}{N} \sum_{i=1}^{N}1(n_{i}^{anode} = \widehat{n}_{i}^{anode}),
\end{equation} 
\begin{equation}\label{equ:4}
\centering
\small
Acc_{num}^{cathode}=\frac{1}{N} \sum_{i=1}^{N}1(n_{i}^{cathode} = \widehat{n}_{i}^{cathode}),
\end{equation} 
\begin{equation}\label{equ:5}
\centering
\small
Acc_{num}^{pair}=\frac{1}{N} \sum_{i=1}^{N}1(n_{i}^{pair} = \widehat{n}_{i}^{pair}),
\end{equation} 
\begin{equation}\label{equ:6}
\centering
\small
\begin{split}
MAE_{position}^{anode}=\frac{1}{N} \sum_{i=1}^N\Biggl(\frac{1}{n_i^{anode}} \sum_{j=1}^{n_i^{anode}}\left|p_{i, j}^{anode}-\widehat{p}_{i, j}^{anode}\right|\Biggr),
\end{split}
\end{equation} 
\begin{equation}\label{equ:7}
\centering
\small
\begin{split}
MAE_{position}^{cathode}=\frac{1}{N} \sum_{i=1}^N\Biggl(\frac{1}{n_i^{cathode}} \sum_{j=1}^{n_i^{cathode}}\left|p_{i, j}^{cathode}-\widehat{p}_{i, j}^{cathode}\right|\Biggr),
\end{split}
\end{equation} 
\begin{equation}\label{equ:8}
\centering
\small
\begin{split}
MAE_{overhang}^{pair}=\frac{1}{N}\sum_{i=1}^N\Biggl(\frac{1}{n_i^{{cathode}}} \sum_{j=1}^{n_i^{{cathode}}}\Bigl|\bigl(\bigl|p_{i, j}^{{cathode}}-{p}_{i, j}^{anode}\bigr|+
\\
\bigl|p_{i, j}^{{cathode}}-p_{i, j+1}^{anode}\bigr|\bigr)-\bigl(\bigl|\widehat{p}_{i, j}^{cathode}-\widehat{p}_{i, j}^{anode}\bigr|+ \\
\bigl|\widehat{p}_{i, j}^{{cathode}}-\widehat{p}_{i, j+1}^{anode}\bigr|\bigr)\Bigr|\Biggr),
\end{split}
\end{equation} 
where $N$ is the scale of the  test dataset, $n_i^{x}$ and $\widehat{n}_{i}^{x}$ separately represent the number of predicted plates and ground truth plates for the $i^{th}$ sample ($x \in\left \{anode, cathode, pair \right \}$), $p_{i, j}^{{y}}$ and $\widehat{p}_{i,j}^{y}$  separately represent the position coordinates of the $j^{th}$ predicted plate and ground truth plate for the $i^{th}$ sample ($y \in\left \{anode, cathode \right \}$). 
We need to sort the coordinates of the plates before calculating $MAE_{position}^{anode}$, $MAE_{position}^{cathode}$ and $MAE_{overhang}^{pair}$.

Generally, the corresponding $n_i^{x}$ can be obtained by counting the number of $p_{i, j}^{{y}}$. Corner detection methods directly predict the coordinate value of each endpoint. General/Tiny object detection methods predict the bounding box for each endpoint, and we utilize the box to calculate the center coordinates as $p_{i, j}^{{y}}$. Counting methods obtain the $n_i^{x}$ by accumulating the spatial values of the predicted density maps. Our segmentation-based MDCNet can predict each point map, and we obtain the coordinate of the endpoint by calculating the center coordinates for the circumscribed rectangle of each point map.

\begin{table}
	\centering
	\scriptsize
	\resizebox{\linewidth}{!}{
		\setlength\tabcolsep{1pt}
		\renewcommand\arraystretch{1.2}
		\begin{tabular}{|c||c|c|c|c|c|c|c|c|}
  \hline
  Method & AN- & CN- & AN- & CN- & PN- & AL- & CL- & OH-      \\
    & MAE$\downarrow$ & MAE$\downarrow$ & ACC$\uparrow$ & ACC$\uparrow$ & ACC$\uparrow$ & MAE$\downarrow$ & MAE$\downarrow$ & MAE$\downarrow$      \\
  \hline \hline      
Baseline
&2.3892&1.4782&	0.5363	&0.5741&	0.5141&	3.8171	&3.0830&	3.3911
 \\
 + PFM
&1.2320	&0.7389	&0.6467&	0.7035	&0.6340	&3.5168	&2.8312&	3.1730
\\
 + CP
&0.8342	&0.3782	&0.6971	&0.7981	&0.6909	&3.4143	&2.7320	&3.0311
\\
 \rowcolor{mygray}
 + LP
&0.4259	&0.2050	&0.7855	&0.9306	&0.7603	&2.4397	&2.0978	&2.1092
\\
  \hline
\end{tabular}

	}
	\setlength{\abovecaptionskip}{2pt}
	
	\caption{Ablation experiments of each component in the MDCNet on the difficult split.}
	\label{tab:ablation_study_1_difficult}
\end{table}

\begin{table}
	\centering
	\scriptsize

	\resizebox{\linewidth}{!}{
		\setlength\tabcolsep{1pt}
		\renewcommand\arraystretch{1.2}
		\begin{tabular}{|c||c|c|c|c|c|c|c|c|}
  \hline
  Method & AN- & CN- & AN- & CN- & PN- & AL- & CL- & OH-      \\
    & MAE$\downarrow$ & MAE$\downarrow$ & ACC$\uparrow$ & ACC$\uparrow$ & ACC$\uparrow$ & MAE$\downarrow$ & MAE$\downarrow$ & MAE$\downarrow$      \\
  \hline \hline      
Baseline 
&4.3223	&3.3240	&0.2989	&0.4138	&0.2586	&3.7476	&3.1356	&3.6235
 \\
 + PFM
&3.1349	&2.6393	&0.4253	&0.5172	&0.3793	&3.6354	&3.0325	&3.5012
\\
 + CP
&2.4763	&2.1223	&0.4885	&0.6034	&0.4483	&3.6255	&2.9222	&3.3521
\\
 \rowcolor{mygray}
 + LP
&2.0920	&1.8966	&0.5632	&0.6839	&0.5115	&2.0004	&1.4654	&1.6291
\\
  \hline
\end{tabular}

	}
	\setlength{\abovecaptionskip}{2pt}
	
	\caption{Ablation experiments of each component in the MDCNet on the tough split.}
	\label{tab:ablation_study_1_tough}
\end{table}

\begin{table}
	\centering
	\scriptsize
	\resizebox{\linewidth}{!}{
		\setlength\tabcolsep{1pt}
		\renewcommand\arraystretch{1.2}
		\begin{tabular}{|c||c|c|c|c|c|c|c|c|}
  \hline     
  
  Method & AN- & CN- & AN- & CN- & PN- & AL- & CL- & OH-      \\               & MAE$\downarrow$ & MAE$\downarrow$ & ACC$\uparrow$ & ACC$\uparrow$ & ACC$\uparrow$ & MAE$\downarrow$ & MAE$\downarrow$ & MAE$\downarrow$      \\
  \hline \hline      
Const-1
&3.1556	&2.4758	&0.4101	&0.4479	&0.3722 &2.5875	&2.5312	&2.4578

              \\
Const-3
&2.4310	&1.7563	&0.6467	&0.6940	&0.5868	&2.5461	&2.4277	&2.4006

 \\
Const-5
&2.4315	&1.7563	&0.6309	&0.6940	&0.5836	&2.5565	&2.2899	&2.3973

\\
Ada-0.1
&2.8562	&2.0223	&0.5521	&0.6625	&0.4890	&2.4435	&2.1544	&2.2654

\\
 \rowcolor{mygray}
Ada-0.3
&0.4259	&0.2050	&0.7855	&0.9306	&0.7603	&2.4397	&2.0978	&2.1092
\\
Ada-0.5
&0.9785	&0.6876	&0.7035	&0.8486	&0.6688	&2.5851	&2.2954	&2.3025

\\
  \hline
\end{tabular}

	}
	\setlength{\abovecaptionskip}{2pt}
	
	\caption{Quantitative comparison  of different label generation strategies and settings for the point segmentation branch on the difficult split.}
	\label{tab:ablation_study_2_difficult}
\end{table}

\begin{table}
	\centering
	\scriptsize
	\resizebox{\linewidth}{!}{
		\setlength\tabcolsep{1pt}
		\renewcommand\arraystretch{1.2}
		\begin{tabular}{|c||c|c|c|c|c|c|c|c|}
  \hline      
  
  Method & AN- & CN- & AN- & CN- & PN- & AL- & CL- & OH-      \\               & MAE$\downarrow$ & MAE$\downarrow$ & ACC$\uparrow$ & ACC$\uparrow$ & ACC$\uparrow$ & MAE$\downarrow$ & MAE$\downarrow$ & MAE$\downarrow$      \\
  \hline \hline      
Const-1
&6.5420	&5.4246	&0.1724	&0.2356	&0.1207	&2.5854	&1.9462	&2.0996

              \\
Const-3
&4.0102	&3.4432	&0.4195&	0.4943	&0.3621	&2.3844&	1.7215	&1.9235

 \\
Const-5
&4.0674	&3.4432	&0.4023	&0.4943	&0.3506	&2.3457	&1.6947	&1.9021

\\
Ada-0.1
&5.4123	&4.0215	&0.2874&	0.3736	&0.2586&	2.3526&	1.6785	&1.8645

\\
 \rowcolor{mygray}
Ada-0.3
&2.0920	&1.8966	&0.5632	&0.6839	&0.5115	&2.0004	&1.4654	&1.6291
\\
Ada-0.5
&2.2147	&2.0456	&0.4828	&0.5402	&0.4023	&2.1756	&1.5455	&1.7442

\\
  \hline
\end{tabular}

	}
	\setlength{\abovecaptionskip}{2pt}
	
	\caption{Quantitative comparison of different label generation strategies and settings for the point segmentation branch on the tough split .}
	\label{tab:ablation_study_2_tough}
\end{table}

\begin{table}
	\centering
	\scriptsize
	\resizebox{\linewidth}{!}{
		\setlength\tabcolsep{1pt}
		\renewcommand\arraystretch{1.2}
		\begin{tabular}{|c||c|c|c|c|c|c|c|c|}
  \hline        
  
  No. & AN- & CN- & AN- & CN- & PN- & AL- & CL- & OH-      \\               & MAE$\downarrow$ & MAE$\downarrow$ & ACC$\uparrow$ & ACC$\uparrow$ & ACC$\uparrow$ & MAE$\downarrow$ & MAE$\downarrow$ & MAE$\downarrow$      \\
  \hline \hline      
1
&0.4259&	0.2050&	0.7855&	0.9306&	0.7603&	2.4397	&2.0978	&2.1092

              \\
2
&0.4259&	0.2050&	0.7855&	0.9306&	0.7603&	2.4397	&2.0978	&2.1092
 \\
3
&0.4259&	0.2050&	0.7855&	0.9306&	0.7603&	2.4397	&2.0978	&2.1092

\\
4
&0.4259&	0.2050&	0.7855&	0.9306&	0.7603&	2.4397	&2.0978	&2.1092

\\
5
&0.4259&	0.2050&	0.7855&	0.9306&	0.7603&	2.4397	&2.0978	&2.1092

\\
  \hline
\end{tabular}

	}
	\setlength{\abovecaptionskip}{2pt}
	\caption{ Evaluation of the PFM with different prompt inputs on the difficult split.}
	\label{tab:ablation_study_4_difficult}
\end{table}

\begin{table}
	\centering
	\scriptsize

	\resizebox{\linewidth}{!}{
		\setlength\tabcolsep{1pt}
		\renewcommand\arraystretch{1.2}
		\begin{tabular}{|c||c|c|c|c|c|c|c|c|}
  \hline    
  
  No. & AN- & CN- & AN- & CN- & PN- & AL- & CL- & OH-      \\               & MAE$\downarrow$ & MAE$\downarrow$ & ACC$\uparrow$ & ACC$\uparrow$ & ACC$\uparrow$ & MAE$\downarrow$ & MAE$\downarrow$ & MAE$\downarrow$      \\
  \hline \hline      
1
&2.0920	&1.8966	&0.5632	&0.6839	&0.5115	&2.0004	&1.4654	&1.6291
              \\
2
&2.0920	&1.8966	&0.5632	&0.6839	&0.5115	&2.0004	&1.4654	&1.6291
 \\
3
&2.0920	&1.8966	&0.5632	&0.6839	&0.5115	&2.0004	&1.4654	&1.6291

\\
4
&2.0920	&1.8966	&0.5632	&0.6839	&0.5115	&2.0004	&1.4654	&1.6291

\\
5
&2.0920	&1.8966	&0.5632	&0.6839	&0.5115	&2.0004	&1.4654	&1.6291

\\
  \hline
\end{tabular}

	}
	\setlength{\abovecaptionskip}{2pt}
	\caption{ Evaluation of the PFM with different prompt inputs on the tough split.}
	\label{tab:ablation_study_4_tough}
\end{table}

\subsection{Ablation Study on the Difficult and Tough Splits}
\cref{tab:ablation_study_1_difficult} - \cref{tab:ablation_study_4_tough} list the results of ablation study on the difficult and tough splits, which can show the effectiveness of each component in the multi-dimensional collaborative network, label generation strategies and the strong stability of the prompt filter module.

{
    \small
    \bibliographystyle{ieeenat_fullname}
    \bibliography{main}
}

\end{document}